\documentclass[11pt,a4paper]{article}
\usepackage{times, latexsym}
\usepackage{amssymb}
\usepackage{CJKutf8}
\usepackage{multirow}
\usepackage{url}
\usepackage[T1]{fontenc}

\usepackage{hyperref}
\usepackage{microtype}
\usepackage{amsmath}
\usepackage{cleveref}
\usepackage{appendix}
\usepackage[dvipsnames]{xcolor}
\usepackage{booktabs}
\usepackage{textcomp}
\usepackage{graphicx}
\usepackage{soul}

\usepackage{todonotes}
\presetkeys{todonotes}{inline}{}

\usepackage[acceptedWithA]{tacl2018v2}

\newcommand{\mkqa}{MKQA}
\newcommand{\gnq}{NQ}
\newcommand{\respace}{\vspace*{-.1cm}}

\title{MKQA: A Linguistically Diverse Benchmark for\\ Multilingual Open Domain Question Answering}

\author{Shayne Longpre \\
  Apple Inc. \\
  \texttt{slongpre@mit.edu}\\\And
  Yi Lu \\
  Apple Inc. \\
  \texttt{ylu7@apple.com} \\\And
  Joachim Daiber \\
  Apple Inc. \\
  \texttt{jodaiber@apple.com} \\
  }

\date{}

\begin{document}
\maketitle

\begin{abstract}
Progress in cross-lingual modeling depends on challenging, realistic, and diverse evaluation sets. 
We introduce Multilingual Knowledge Questions and Answers (\mkqa{}), an open-domain question answering evaluation set comprising 10k question-answer pairs aligned across 26 typologically diverse languages (260k question-answer pairs in total).
Answers are based on heavily curated, language-independent data representation, making results comparable across languages and independent of language-specific passages.
With 26 languages, this dataset supplies the widest range of languages to-date for evaluating question answering.
We benchmark a variety of state-of-the-art methods and baselines for generative and extractive question answering, trained on Natural Questions, in zero shot and translation settings.
Results indicate this dataset is challenging even in English, but especially in low-resource languages.\footnote{MKQA data and evaluation scripts are available at \url{https://github.com/apple/ml-mkqa}.}
\end{abstract}

\section{Introduction}
\respace
Training and evaluation data for question answering (QA) is severely lacking outside of high-resource languages like English.
As unsupervised, transfer learning and zero/few-shot methods narrow the multilingual performance gap with English \citep{conneau2019unsupervised, lee2019cross, cui2019cross, lewis2019mlqa}, their real progress is hard to measure without challenging, realistic, and linguistically diverse evaluation sets.
Existing multilingual QA datasets are realistic and challenging, but they lack linguistic diversity, comparable evaluation between languages, and are often limited to passages provided with the dataset (see Table~\ref{evalset-comparison}).

We introduce Multilingual Knowledge Questions and Answers (\mkqa{}) for evaluation of open-domain question answering. 
\mkqa{} selects 10k realistic English queries from the Natural Questions dataset \citep[\gnq{},][]{kwiatkowski2019natural} and human translates them into $25$ additional languages and dialects.
Accompanying these query translations we replace \gnq{}'s passage embedded answer spans with high-quality, language- and retrieval-independent answer annotations, linked directly against Wikidata entities and a limited set of well-defined value types (numbers, dates, strings, etc.).\footnote{Wikidata is a collaboratively edited open knowledge graph: \url{https://www.wikidata.org/}.}

See one full example in Table~\ref{tbl:answer-examples}.
More flexible than existing multilingual datasets, \mkqa{}'s grading procedure ensures these labels are sufficient to evaluate any QA method, including knowledge graph and generative approaches.
The objective of this evaluation set is to facilitate fair comparison between languages, without imposing assumptions on the underlying QA approach.
We see \mkqa{} as a useful tool enabling practitioners to benchmark a variety of multilingual open domain question answering methods against the widest range of available languages yet.
Below, we discuss its central properties as an evaluation benchmark.

\begin{table*}
\centering
\includegraphics[width=\linewidth]{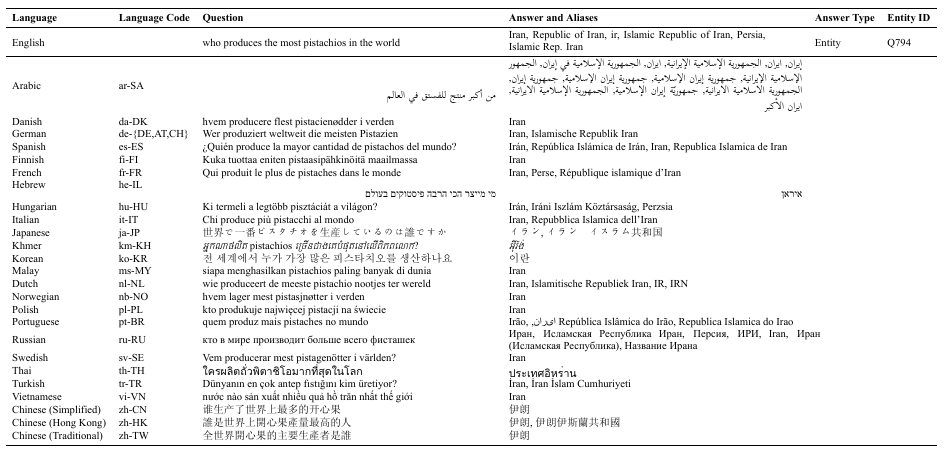}
\caption{\textbf{Questions and answers in all supported languages for one instance in MKQA.} The IETF BCP-47 language codes specify the language and locale. 
The Entity ID corresponds to Wikidata (see for instance \url{https://www.wikidata.org/wiki/Q794}).
}
\label{tbl:answer-examples}
\vspace*{-.4cm}
\end{table*}

\respace
\paragraph{Realistic \& Reliable Annotations}
Of crucial importance to any evaluation set is (a) how well it reflects realistic, real-world settings, and (b) the reliability of its annotations. 
To ensure the English queries, which form the basis of our dataset, are realistic, we use Natural Questions, formulated by real users, independent of passages or answers.
To ensure these queries are realistic in other languages we employ expert bilingual translators, guided by strict localization criteria. 
We confirm that a large majority of these queries are \textbf{geographically invariant}, meaning that their answer is not culturally or geographically dependent (we found that less than 4\% of answers are rendered incorrect by geographical and cultural context, for more details see Section ~\ref{sec:annotationquality}). 
To ensure annotation reliability, we enforce minimum inter-grader agreement, conduct quality checks, and re-annotation from expert graders where necessary.
Further, the Wikidata entity identifiers (QIDs) ground the answer annotations in structured data.
This can be used for other knowledge graph-specific metrics, to retrieve other valid answer strings, and trivial entity translation into hundreds of languages beyond the scope of \mkqa{}.

\respace
\paragraph{Parallel Questions}
Our evaluation set is fully aligned, or ``parallel", across all available languages, meaning the same examples exist in all languages.
This is accomplished by a mixture of expert human translation and using multilingual data from Wikidata.
This property enables direct comparison between all 26 languages for fully cross-lingual or zero-shot systems.
While \citet{clark2020tydi} point out the natural query distribution varies by language and geography, we reserve our assessment to geographically invariant queries for the purpose of more fair comparison between methods.

\respace
\paragraph{Retrieval-Independent Annotations}
Existing training and evaluation sets are oriented to ``extractive'' QA, providing specific passages and passage-dependent answer annotations \citep{clark2020tydi, lewis2019mlqa, artetxe2019cross, liu2019xqa}.
These types of annotations are of limited use with varying retrieval systems, knowledge graph approaches, and even generative approaches because the answers are tied to the particular phrasing of their passage.
Translating annotations from English passages may also introduce ``translationese artifacts" as the translation is implicitly influenced by the original English structure \citep{artetxe2020translation}.
These artifacts render the task easier for methods relying on English supervision or machine translation techniques.
As we shall discuss in \Cref{collection}, the \mkqa{} collection procedure yields primarily entity and structured ``atomic" answer types.
We contend retrieval-independent (and particularly entity-oriented) annotations minimize the risk of translation artifacts, and remove limitations on the underlying QA approach.

\begin{table*}
\centering
\small
\resizebox{1\linewidth}{!}{
\begin{tabular}{lccccc}
\toprule
\textbf{\begin{tabular}{@{}c@{}}Multilingual QA \\ Evaluation Set\end{tabular}} & \textbf{\begin{tabular}{@{}c@{}}Answer \\ Independence\end{tabular}} &
\textbf{\begin{tabular}{@{}c@{}}Parallel \\ Questions\end{tabular}} & 
\textbf{\begin{tabular}{@{}c@{}}Language Fam. \\ Branches\end{tabular}} &
\textbf{Languages} & 
\textbf{Total Examples} \\
\midrule
XQA \citep{liu2019xqa} & \checkmark & ${\times}$ & 5 & 9 &  $28k$ \\
MLQA \citep{lewis2019mlqa} & ${\times}$ & \checkmark & 6 & 7 & $46k$ \\
XQuAD \citep{artetxe2019cross} & ${\times}$ & \checkmark & 11 & 11 & $13k$ \\
TyDi \citep{clark2020tydi} & ${\times}$ & ${\times}$ & 11 & 11 & $204k$ \\
Xor-QA \citep{asai2020xor} & ${\times}$ & ${\times}$ & 7 & 7 & $40k$ \\
\midrule
\mkqa{} (This work) & \checkmark & \checkmark & 14 & 26 & $260k$ \\
\bottomrule
\end{tabular}
}
\caption{\label{evalset-comparison}
\textbf{Comparison of multilingual QA evaluation sets.} Answer independence indicates whether the gold answer is independent of a retrieved document, and parallel questions indicates whether examples are the same across languages.
}
\vspace*{-.4cm}
\end{table*}

\respace
\paragraph{Linguistic Diversity}
Lastly, \mkqa{} has broad linguistic diversity, covering 26 languages and dialects from 14 language family branches.
Languages from \mkqa{} cover half of the world populations' native language, and more than 90\% of the world population lives in a country where one of these languages is an official language (see Section~\ref{sec:languageselection} for more details).
It is to our knowledge both the largest and most linguistically diverse open-domain QA evaluation set currently available (see Tables~\ref{evalset-comparison} and \ref{tbl:languages}).

\respace
\paragraph{}
\mkqa{} makes two important contributions to the field of multilingual question answering:
\begin{itemize}\itemsep0em
\item  Our \textit{answer collection procedure} renders the evaluation set highly reliable, independent, and unbiased towards the QA technique used. 
This unique setup allows us to fairly compare the performance of techniques as distinct as knowledge graph-based, dense and sparse retrieval and generative QA techniques on a large number of languages (see \Cref{experiments}).
\item Our dataset provides \textit{fully aligned} examples in the largest yet number of typologically diverse languages, enabling comparable evaluation across many languages.
\end{itemize}
\respace

We find \mkqa{} is innately more challenging than Natural Questions from which it was derived, due to the multi-stage re-annotation process.
The best model obtains only 52.3\% F1 in English, and only 5.7\% above a naive baseline on the lowest resource language. 
Given these qualities, our dataset facilitates broad and reliable evaluation of multilingual, open-domain question answering.

\respace
\section{Related Work}
\respace
\label{related-work}
\respace
\paragraph{Cross-Lingual Modeling}
Recent work trains cross-lingual representations with unsupervised language modeling over many languages, including Multilingual BERT \citep{devlin2019bert}, XLM-R \citep{conneau2019unsupervised}, and Multilingual T5 \citep{xue2020mt5}.
Transfer learning techniques are often applied to these cross-lingual representations to overcome the dearth of non-English data \citep{cui2019cross, hsu-etal-2019-zero, lee2019cross, kumar2019cross}.
Recent investigations into cross-lingual modeling have revealed ``translation artifacts" in datasets where machine translation systems are used, or human translation tasks are not carefully curated \citep{artetxe2020translation, wintner-2016-translationese, rabinovich2015unsupervised}.
``Translationese" results in hidden linguistic cues in translated text that render the task easier than a natural translation.

\respace
\paragraph{English QA Resources}
A majority of question answering research focuses on English, which offers ample selection of evaluation datasets, including SQuAD \citep{rajpurkar2016squad}, TriviaQA \citep{joshi2017triviaqa}, and Natural Questions \citep{kwiatkowski2019natural}.
Open Domain QA, pioneered by \citet{10.5555/21922.24354}, is the task of answering open questions using external knowledge sources.
A common approach is to combine retrieval and extractive techniques \citep{chen-etal-2016-thorough, chen-etal-2017-reading, dhingra-etal-2017-gated, cui-etal-2017-attention}.

\respace
\paragraph{Monolingual QA Resources}
Non-English question answering resource options remain comparatively rare, with most options spanning only one other language, and rarely low-resource languages.
DuReader \citep{he2018dureader}, CMRC \citep{cui2019span}, and DRCD \citep{shao2018drcd} all offer high-quality Chinese QA datsets.
Similarly, XCMRC \citep{liu2019xcmrc} and BiPar \citep{jing2019bipar} present parallel, cross-lingual QA dataset between English and Chinese.
Exploring slightly less resource-rich languages, numerous works have derived new datasets from SQuAD, employing varying degrees of human or semi-automatic translation techniques to non-English target languages: ARCD for Arabic \citep{mozannar2019neural}, KorQuAD-1.0 for Korean \citep{lim2019korquad1}, and MMQA for Hindi \citep{gupta2018mmqa}.

\respace
\paragraph{Multilingual QA Resources}
Table~\ref{evalset-comparison} compares the largest publicly available multilingual question answering evaluation sets.
The table highlights the following properties of each dataset: whether the available gold answers are independent of retrieved documents, whether examples are aligned across languages, and the number of languages and examples provided.
MLQA \citep{lewis2019mlqa} and XQuAD \citep{artetxe2019cross} are examples of SQuAD-style extractive datasets, employing human translators to create parallel examples.
Both MLQA and XQuAD ensure that all answers are answerable (discarding ``No Answer'' examples), and derive answers from provided documents.
XQA \citep{liu2019xqa}, one of the few retrieval-independent QA datasets, offers cloze-style questions, leveraging Wikipedia's daily questions and entity answers to populate document-independent answers.
TyDi \citep{clark2020tydi}, like \mkqa{}, focuses on typological diversity in its wide language selection.
While TyDi offers a more natural distribution of questions, its annotations are based on the retrieval system used by the authors (Google search); hence their answers are actually start and end indices for spans of text within a given passage.
Xor-QA \citep{asai2020xor} explores cross-lingual subtasks by re-annotating 40k TyDi examples, over 7 languages, sourcing answers from English documents and translating them back to the target language.
Many of these multilingual resources have been bundled into cross-lingual benchmarks, such as XTREME \citep{hu2020xtreme} and XGLUE \citep{liang2020xglue}.

\respace
\subsection{Comparison to Native Speaker Datasets}
\label{sec:native-datasets-comparison}
\respace

There are key advantages to datasets such as TyDi \citep{clark2020tydi} and Xor-QA \citep{asai2020xor} which use native speakers questions, particularly in the naturalness and cultural authenticity of the corpora.
However, there are also key disadvantages to these datasets which \mkqa{} circumvents with language alignment, to provide more challenging and fair model evaluations across languages.

TyDi \citep{clark2020tydi} and \mkqa{} both target high typological diversity, highlight the importance of sourcing realistic questions (with answers unseen), and incorporate a broader distribution of question types than competing datasets (including ``No Answer'' and ``Yes''/``No'' answers).
There are three main differences between \mkqa{} and TyDi: (a) question alignment across languages, (b) answer distribution, and (c) annotation retrieval independence (closely tied with the notions of ``open" and ``closed" domain).
TyDi provides a different set of natural questions per language, at the expense of direct comparability across languages.
Not only are the TyDi questions different between languages, but the percentage of answerable passages varies dramatically, from $22\%$ in Korean to $69\%$ in Arabic.
XorQA-TyDi \citep{asai2020xor} partially resolves this issue by sourcing answers from English documents, but this may in turn re-introduce cultural biases.
This suggests that the conceptual difficulty of these questions may also vary dramatically, as consumers from different locales cater their questions based on their existing beliefs of the quality of the virtual assistants in their language.
As a result, it is difficult to interpret the core reasons why multilingual system's performance varies between languages. 
To ensure this property, \mkqa{} verifies its questions are predominantly geographically invariant, and thus the answers will not change due to geographical or cultural factors.

The second difference between datasets is the answer distribution.
\mkqa{} answers (a) are predominantly entities ($42.2\%$) or atomic answers such as dates, binary, or numbers with units, and (b) use a different definition of ``Unanswerable".
Xor-QA focuses only on answerable queries, TyDi's definition conditions on the presence of the answer in the passage, whereas \mkqa{}'s definition is based on the ability of a human to find a succinct answer to a question on the web, i.e.\ whether it is human answerable.
As a result, our annotations are not limited by the quality of selected passages, and provide higher answer coverage ($67.58\%$ as opposed to the TyDi language average of $38\%$).

Finally, while \mkqa{} does not expect an answer to be derived from a single source document, TyDi is an extractive QA dataset.
Consequently, its answer annotations are defined as spans, tied directly to particular Wikipedia documents and fixed index from which they were retrieved.
As an evaluation set we contend the flexibility of document-independent answers is critical to not restrain what approaches can be evaluated in future research.

\respace
\section{Dataset Collection}
\respace
\label{collection}
\begin{figure*}
\centering
\includegraphics[width=\linewidth]{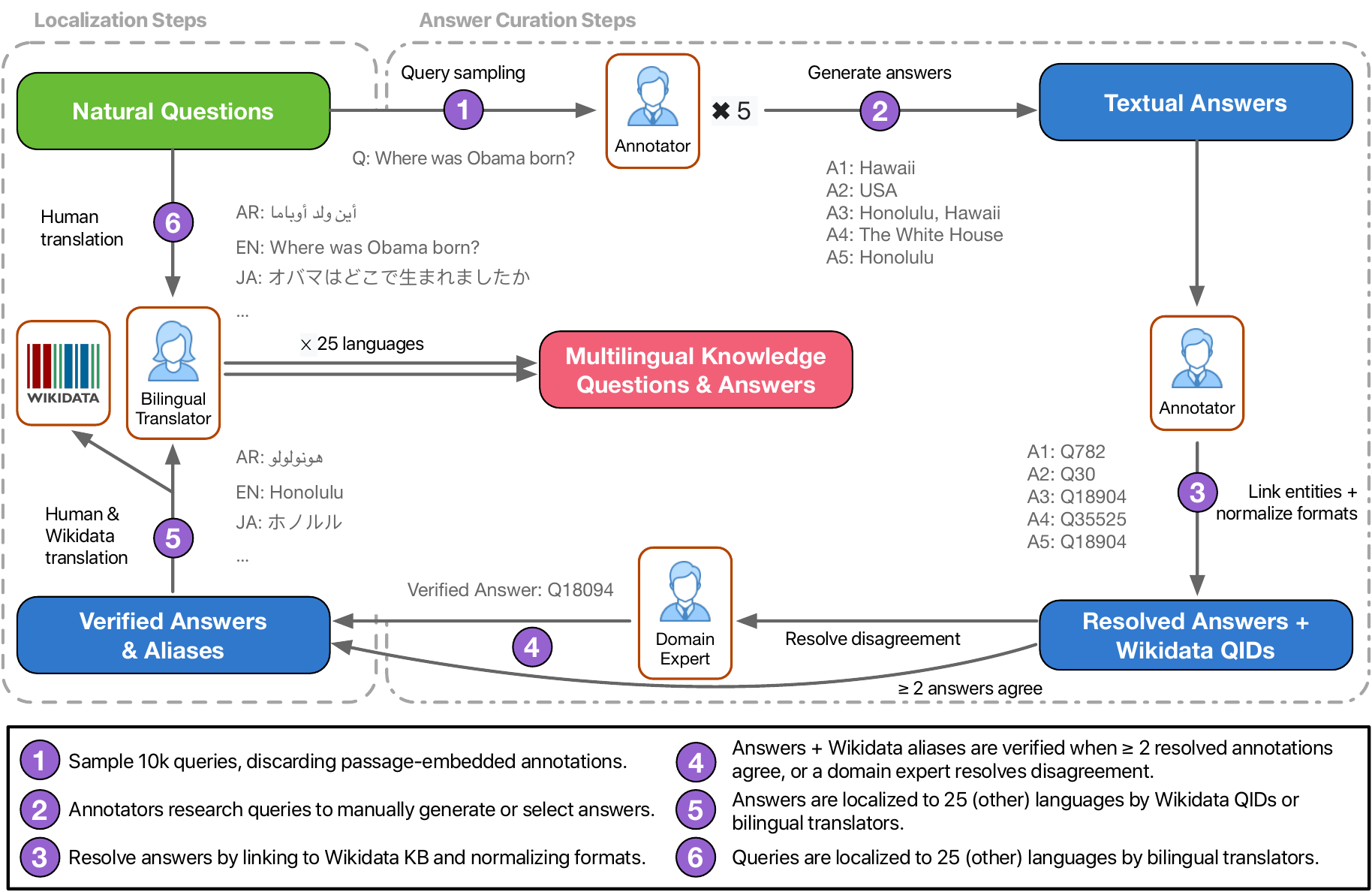}
\caption{\textbf{Data Collection Process.} A depiction of the 6 sequential steps in our data collection pipeline. The first four steps involve Answer Curation, and the last two localize questions and answers into 26 target languages.}
\label{fig:data-collection}
\vspace*{-.4cm}
\end{figure*}

We aim for certain properties of our evaluation set: (i) realistic questions, (ii) reliable annotations (e.g.\ via inter-annotator agreement), and (iii) a flexible task setup that makes as few assumptions as possible about the underlying modeling techniques, enabling fair comparison between any approach.

\respace
\subsection{Query Selection}
\respace

Our evaluation set collection pipeline begins with the Answer Curation steps outlined in Figure~\ref{fig:data-collection}.
These are designed to yield high-concensus answer labels, with normalized textual formats, expressive alias sets for robust comparison, and grounding in structured information for entity disambiguation or more informative analysis.
For the first step, we sample 10,000 queries from Natural Questions (\gnq{}) \citep{kwiatkowski2019natural}, as this is one of the few QA datasets based on realistic queries, generated by information seeking users.

\respace
\subsection{Raw Answer Collection}
\respace

At the raw answer collection stage, 5 annotators are independently shown the query and asked to search the web to either copy \textit{or generate} an ideal answer.
They are asked to select an answer type (radio buttons) from the options shown below, and input the answer (text box) according to format instructions per answer type.
The formatting constraints allow us to automatically link WikiData entities for the units in ``number with units" and to gather well-structured data for answers such as dates, to save annotator time.

\respace
For each query, the graders select a typed answer from the following taxonomy:
\begin{itemize}\itemsep0em
  \item \textbf{Atomic value:} This category includes dates, numbers and number ranges with or without a unit (meters, years, ...).
  \item \textbf{Entities:} Entities are annotated with Wikidata QIDs and include generic entities, people, objects, and most locations.
  \item \textbf{Yes/No:} Type representing yes/no answers.
  \item \textbf{Short answer:} Answers which cannot be encapsulated in an atomic value, entity or binary (yes/no) answer, but are still a short phrase.
  \item \textbf{Long answer:} The long answer category indicates no simple factual answer or short phrase answers this question and a longer or visual explanation is required. During evaluation we treat these as ``Unanswerable" for simplicity.
  \item \textbf{Unanswerable:} This category indicates that the query is not answerable, potentially because it is ill-formed or because no clear answer is available.
\end{itemize}
\respace

\subsection{Answer Resolution}
\respace

Given the query and a candidate answer from the previous stage, annotators are next asked to normalize date/number formats and resolve the answer text against Wikidata entities, where feasible.
To resolve short textual answers against Wikidata entities, we apply an internal entity linking system to the answer string to generate Wikidata candidate entities.\footnote{This step can be replicated using an off-the-shelf entity linker such as spaCy available at \url{https://spacy.io/api/entitylinker}.} The top 10 entity suggestions and their descriptions, along with the original query and short answer are then presented to 3 graders, who are asked to pick the correct reference entity or ``None of the above.'' In cases where graders do not achieve sufficient agreement or where the correct entity is not in the list, a domain expert (one of the MKQA authors/designers) provides the correct reference.
Overall, this step enables us to disambiguate homonyms and collect valid answer synonyms/aliases, for more robustly measuring annotator agreement and prediction accuracy.
   
\respace 
\subsection{Answer Verification}
\respace

Up until this stage, 5 raw answers were collected per query, and subsequently format normalized and resolved against Wikidata.
In the fourth stage of Answer Curation (in Figure~\ref{fig:data-collection}) any normalized answer given by at least 2 annotators is admitted to the final set as a gold answer.
For those annotations that did not achieve the required agreement from at least two annotators, a domain expert (one of the \mkqa{} authors/designers) with access to all 5 preliminary annotations is tasked to provide a final decision.
This second manual round was afforded as much time per decision as necessary to obtain a satisfactory answer.
The instructions permit the selection of existing normalized answer(s), modifying them slightly, or overriding them if necessary.

\respace
\subsection{Answer Localization}
\respace

In the last two stages of \mkqa{} curation shown in Figure~\ref{fig:data-collection} we translate, or ``localize" the English queries and answers into the target languages.
Given the special care we took to avoid them in our methodology, and since we only localize short answers and queries (no context passages), we believe translation artifacts are likely to be minimal in \mkqa{}.

Verified answers are localized into the target language by a combination of methods.
For Wikidata-resolved answers, we leverage Wikidata's names and aliases for the target language.
These names and aliases are transcribed in the native alphabet where appropriate, reflecting the expected answer in each language.
Atomic answer types, including numeric, number with entity, and date types were also translated by this method, maintaining Arabic numerals for all languages, but naturalizing unit terms such as ``November", ``century", ``b.c", ``acres", and ``light years".
For date types specifically, for every combination of year, month, and day, we generate template answers in each language, accommodating both American and European date formats, as well as numeric and written out versions for months.

In cases where a Wikidata link could not be found, or where answers were not available for a given language code, professional bilingual human translators were used to provide the native equivalent.
For this task, human translators are given access to the English query, the English answer, and where available the Wikidata link and Wikipedia page for the entity.
We found localization quality improved when bilingual translators are shown several examples prior to grading, covering each of the localization options:

\respace
\paragraph{Localization Options:}
\begin{itemize}\itemsep0em
\item \textbf{Transliteration} is a type of conversion of a text from one script to another that involves swapping letters (thus trans- + liter-) in predictable ways (such as $\alpha$ $\rightarrow$ a, $\chi$ $\rightarrow$ ch, or æ $\rightarrow$ ae). 
\item \textbf{Translation} is the communication of the meaning of a source-language text by means of an equivalent target-language text.
\item \textbf{Unchanged} is selected if the entity name does not need to be localized as it is commonly used as is.
\item \textbf{Mix transliteration/translation/unchanged} if the entity is localized using more than one technique.
\end{itemize}
\respace

\respace
\subsection{Query Localization}
\respace

The final stage of \mkqa{} construction, as shown in Figure~\ref{fig:data-collection}, is query localization.
As with answer localization, bilingual translators were asked to translate each query ensuring the query's meaning is maximally preserved, while naturally phrased.
Translators were further instructed to use localized names of named entities if they exist in the target language and to transliterate names otherwise.
Our translators, who are native speakers of the target language, are verified to live in the targeted region and are required to pass an entrance exam to verify a high level of fluency in English.
Translators received a standard hourly wage varying with the target region and were not compensated per completed task, as is usual with alternative public services such as Mechanical Turk.
On average, around 16 translators participated in the translation of the 10k source queries from English into each target language.

\respace
\section{Dataset Quality and Analysis}
\respace
\label{data-quality}
Given our dataset collection and methodology, we evaluate the effect of our choices, and the properties of the final set, including the selected languages, annotation quality, geographical invariance, and answer type distribution as compared to \gnq{}.

\respace
\subsection{Language Selection}
\label{sec:languageselection}
\respace

\begin{table}[t]
\begin{center}
\small
\resizebox{1\linewidth}{!}{
\begin{tabular}{llll}
\toprule
\textbf{Family}                 & \textbf{Branch}               & \textbf{Language}  & \textbf{Reach} \\ 
\toprule
\multirow{12}{*}{Indo-European} & \multirow{6}{*}{Germanic}     & English               & 16.46\%                                      \\
                                &                               & German             & \phantom{0}1.70\%                                       \\
                                &                               & Dutch              & \phantom{0}0.38\%                                       \\
                                &                               & Swedish            & \phantom{0}0.17\%                                       \\
                                &                               & Danish             & \phantom{0}0.08\%                                       \\
                                &                               & Norwegian          & \phantom{0}0.07\%                                       \\\cmidrule(l){2-4} 
                                & \multirow{4}{*}{Italic}       & Spanish            & \phantom{0}6.99\%                                       \\
                                &                               & French             & \phantom{0}3.59\%                                       \\
                                &                               & Portuguese         & \phantom{0}3.28\%                                       \\
                                &                               & Italian            & \phantom{0}0.87\%                                       \\\cmidrule(l){2-4} 
                                & \multirow{2}{*}{Balto-Slavic} & Russian            & \phantom{0}3.35\%                                       \\
                                &                               & Polish             & \phantom{0}0.58\%                                       \\
\midrule
\multirow{2}{*}{Sino-Tibetan}   & \multirow{2}{*}{Sinitic}      & Mandarin              & 14.54\%                                      \\
                                &                               & Cantonese             & \phantom{0}1.10\%                                            \\ 
\midrule
\multirow{2}{*}{Afro-Asiatic}   & \multirow{2}{*}{Semitic}      & Arabic                & \phantom{0}4.44\%                                       \\
                                &                               & Hebrew                & \phantom{0}0.12\%                                       \\
\midrule
Austronesian                    & Malayo-Poly.                  & Malay              & \phantom{0}3.47\%                                       \\

\midrule
Japonic                         & Japonic                       & Japanese           & \phantom{0}1.64\%                                       \\
\midrule
\multirow{2}{*}{Austroasiatic}  & Vietic                        & Vietnamese         & \phantom{0}1.00\%                                       \\
                                & Khmer                         & Khmer              & \phantom{0}0.21\%                                       \\
\midrule
Turkic                          & Com.\ Turkic                  & Turkish            & \phantom{0}1.10\%                                       \\
\midrule
Kra–Dai                         & Tai                           & Thai               & \phantom{0}0.78\%                                       \\
\midrule
Koreanic                        & Han                           & Korean             & \phantom{0}1.03\%                                       \\
\midrule
\multirow{2}{*}{Uralic}         & Finnic                        & Finnish            & \phantom{0}0.07\%                                       \\
                                & Ugric                         & Hungarian          & \phantom{0}0.17\%                                       \\
\bottomrule
\end{tabular}
}
\end{center}
\caption{\textbf{Languages with their corresponding language families and speakers.} \textit{Reach} indicates the combined number of first-language (L1) and second-language (L2) speakers as a percentage of the world population \citep[Ethnologue,][]{ethnologue}.}
\label{tbl:languages}
\vspace*{-.4cm}
\end{table}

We select a set of languages meeting both academic and practical considerations, by maximizing typological diversity as well as the share of the world population that understand at least one of the languages in the set.
Table~\ref{tbl:languages} shows the languages selected for our dataset with the corresponding branch of their language family. 
We also show the language's reach, i.e.\ the percentage of the world population that speaks the language either as a first or second language \cite[based on Ethnologue data,][]{ethnologue}.
Since combined first- and second-language speaker statistics are not readily available, it is not straight-forward to accurately determine what share of the world population can be covered by the languages in this set (e.g.\ a native speaker of German may also be fluent in English).
A practical option is to calculate the share of the world population that lives in a country where one of the languages in our set is recognized as an official language.
By this measure, 90.62\% of the world population live in a country with an official language covered by the languages in our set.\footnote{
We determine this percentage based on Wikidata as the combined population (Wikidata property ``P1082'') of all countries that have an official language (Wikidata property ``P37'') in our dataset divided by the combined population of all countries in Wikidata.}
With the large number of diverse language families covered and the reach of the selected languages, \mkqa{} addresses both academic and practical requirements for a wide and diverse question answering benchmark.
Finally, we note that the Wikidata IDs provided for a large portion of our gold answers allow these answers to be further localized into Wikipedia languages beyond those in \mkqa{}, should practitioners wish to expand their analysis.

\begin{figure*}
\centering
\includegraphics[width=\linewidth]{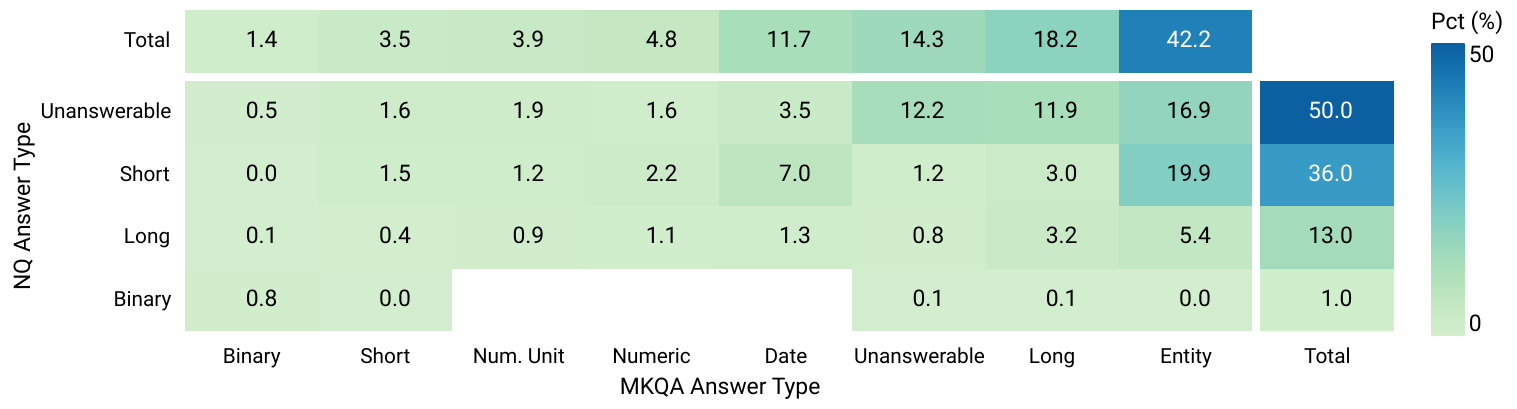}
\caption{\textbf{Answer Type Breakdown.} Compares the distribution of answer types between MKQA and Natural Questions (NQ) for the $10k$ examples in the evaluation set.}
\label{fig:answer-type-breakdown}
\vspace*{-.4cm}
\end{figure*}

\respace
\subsection{Translation and Answer Quality}
\label{sec:annotationquality}
\respace

The quality and reliability of our dataset is highly dependent on two factors: (a) How well our professional translators were able to translate the English queries into each target language, and (b) how well our language-independent answer representations transfer to each target language.

We run a small-scale grading experiment, grading just above 1\% of the total data, to estimate the quality of the query translations and how well the meaning of our language-independent answer annotation is preserved across languages (geographical invariance).
We present graders with the localized query and its answer annotations
and ask them to judge whether (a) the localized query is an acceptable translation of the original English query, and (b) whether the provided answer (entities are shown with their QID and description, and a short explanation is added to each other answer type) is acceptable for the translated target-language query. 
In addition, we also ask graders to judge the answer quality for the original English queries as a baseline.

\begin{table}
\small
\centering
\begin{tabular}{lcc}
\toprule
\multirow{2}{*}{\textbf{Language}}   & \multicolumn{2}{c}{\textbf{Acceptance Rate}} \\
                                     & \textbf{Query Translation} & \textbf{Answer} \\ \midrule
English                        & - &  97.03\%                  \\
\midrule
German                         & 99.01\% &  91.08\%                  \\
Spanish                        & 99.01\% &  92.07\%                  \\
Thai                           & 96.04\% &  91.09\%                  \\
Chinese (simpl.)               & 92.24\% &  89.32\%                  \\
\bottomrule
\end{tabular}

\caption{\textbf{Query translation and retrieval-agnostic answer quality in various languages.} Query translation acceptance rate is the percentage of query translations judged as acceptable. Answer acceptance rates is the percentage of answers graders found acceptable in response to the translated target-language query.}
\label{tbl:answergrading}
\vspace*{-.4cm}
\end{table}

Table~\ref{tbl:answergrading} shows the acceptance rates for query translations and answers for a small selection of languages. The table shows
that query translations are consistently judged as acceptable in German, Spanish and Thai, while the quality for Chinese translations was judged as lower in comparison. 
Most translation issues are related to the localization of entities and to domain-specific terms (e.g.\ sports terminology such as ``receptions'' in football). 
As expected, the acceptability of answers is judged to be higher for English than other languages but it is still at or above 90\% even for languages as linguistically distant from English as Thai.
Note that errors in answer acceptance rate and query translation acceptance rate heavily overlap since incorrect query translations will most likely mean that the existing language-independent answer will not match.
Answer quality issues fall into the following categories (illustrated with German examples):

\textbf{(1) Answer differs based on cultural context (44\%)} This includes cases where the localized version of an entity may have different properties. 
For example the English-language TV show ``man vs food'' has 8 seasons while the German version has 5. 
Similarly a character in a movie such as ``Finding Nemo'' may be voiced by a different voice actor in the German version of the same movie.

\textbf{(2) Generic annotation issues (33\%)} The second biggest source of errors are answer quality issues that will hold across languages. 
Examples include answers that are time-sensitive such as the answer to the question ``when was the oldest person in the world born'' and questions with ambiguous answers in the data such as ``is northern ireland a part of great britain.''

\textbf{(3) Entities transliterated incorrectly (11\%)} Names for entities may be transliterated incorrectly if they do not exist in the target language (``who wrote the book clear and present danger'').

\textbf{(4) Generic translation artifacts (11\%)} Generic translation errors may lead to a mismatch between the question and the language-independent answer. 
In one example the English ``words to'' meaning ``lyrics'' was translated into German as the literal ``Worte'' which would be an uncommon phrasing in a question about lyrics.

Translation artifacts are a recognized problem in multilingual datasets and manual grading of the data in Table~\ref{tbl:answergrading} shows that the human translation step may introduce more or less query--answer discrepancies depending on the target language.
In an alternative scenario, annotation could be performed directly on native queries from each language; however, such data is not readily available and might additionally suffer from other downsides such as relatively small user bases in less frequently spoken languages (see Section~\ref{sec:native-datasets-comparison} for further discussion).
Similar to our evaluation, the authors of \gnq{} perform a manual precision grading of their data and find an overall data precision of 84\% for short answers.
While we hope that future work can improve on data quality further, comparatively even for the language with the most severe translation artifacts in our evaluation, Simplified Chinese, the resulting data quality (answer acceptance rate of 89\%) is still within an acceptable range.
In addition, our dataset provides the only available source of question answering evaluation in many languages.

We encourage authors of future multilingual datasets that use any translation methods to report and detail their geographical invariance, as we have done, and to benchmark the reliability of examples and presence of translation artifacts.

\respace
\subsection{Annotation Breakdowns}
\label{annotation-breakdown}
\respace

Next, we compare the distribution of answer types between the original \gnq{} dataset, with those newly assigned in \mkqa{}. 
As Figure~\ref{fig:answer-type-breakdown} shows, $50\%$ of \gnq{} are completely ``Unanswerable" by retrieved passages and another $13\%$ require long passage answers.
In the short answer setup for \gnq{} both of these are considered unanswerable, amounting to $63\%$ of all questions.
In comparison, only $32.4\%$ of examples are ``Unanswerable" or ``Long" answer type in \mkqa{}.
This is due to a shift in definition from whether a passage contains an answer, to whether a question is (succinctly) answerable by a human, with full web access.
Given that the answer types in \mkqa{} are not dependent on a learned retrieval system, they reflect the properties of the question only.

We later show that this ``unanswerable" definition yields more challenging evaluation because (i) correctly answering questions is on average harder than learning when to abstain, and (ii) many of the most difficult questions were unanswerable in \gnq{} but are answerable in \mkqa{}.
This suggests the property of ``retrieval independent annotations", currently not used in any other multilingual QA benchmarks except XQA, is highly desirable for (a) constructing more challenging QA evaluation sets, and (b) yielding annotations useful to evaluate any QA approach, not just extractive QA models. 

We also encourage future QA benchmarks to mimic our multi-stage data collection framework in providing supplementary metadata per example (answer type and Wikidata QIDs).
Beyond basic comparison of systems, our evaluation tools allow practitioners to perform further error analysis with more interpretable metrics.

\respace
\section{Experiments}
\respace
\label{experiments}
\subsection{Task Definition}

Given a question $q^{l}$ in language $l$, the task is to produce a prediction ${p^{l} \in {\{\text{No Answer, Text Answer}\}}}$, where a Text Answer is a sequence of tokens in the corresponding language. 
$p^{l}$ can be obtained by any method, extracted from a document, generated, or derived from a knowledge graph.

For evaluation using \mkqa{} gold answers, every question $q^{l}_i$ from $i \in [1,10000]$ is accompanied by a set of valid annotations $a^{l}_i$ per language. 
Every prediction $p^{l}_i$ is scored based on exact match (\textsc{EM}) and token overlap \textsc{F1}, as with previous open-retrieval QA datasets. 
The official evaluation script also ingests a ``No Answer probability" for each example.
If the probability is above a chosen threshold value then the prediction defaults to No Answer instead of the provided Textual Answer.
As this threshold varies from 0 to 1 the predictions shift from entirely No Answer to all textual answers.
We follow \gnq{} in reporting the best \textsc{F1} over the range of thresholds, to remove threshold tuning as a factor in evaluation.
A best threshold is computed and applied per language, where each example receives a  ``textual" (token overlap) F1 after language-specific normalization (removing whitespace, punctuation, and articles) is applied to both the prediction and gold answers.
Finally, the official per-language \textsc{F1} is computed as the mean of example F1s, and the official Macro Average \textsc{F1} is the mean of per-language \textsc{F1} scores.

\respace
\subsection{Baseline Approaches}
\label{ext-baselines}
\respace

To benchmark our evaluation set, we combine state-of-the-art approaches in retrieval, machine translation, extractive QA and generative QA.
All retriever models are off-the-shelf, and all reader models are finetuned on Natural Questions, including \textsc{Xlm-Roberta Large} \citep{conneau2019unsupervised} and \textsc{M-Bert} \citep{devlin2019bert} for extractive QA, and \textsc{mT5-Large} \citep{xue2020mt5} for generative QA.\footnote{Note that we exclude the $10k$ examples used in our evaluation set from this training set.}
In each case, tokenization is handled by the multilingual model used --- sentencepiece for \textsc{Xlm-R} and \textsc{mT5-Large}, WordPiece for \textsc{M-Bert}, each with vocabularies initialized from their specific pre-training implementations.
Further, all query and prediction translations in our approaches use \citet{zhang2020improving}'s open source many-to-many, encoder-decoder machine translation system, trained on the OPUS multilingual corpus, covering 100 languages.

\paragraph{Retrieval Corpora} Our baselines operate on a Wikipedia document corpus from Dec. 07, 2020 following previous work in open-domain question answering \citep{kwiatkowski2019natural, asai2020xor, clark2020tydi}.
We use the language-specific Wikipedia corpora for Elasticsearch and the English versions for other baselines. Using Wikipedia as this base corpus is a pragmatic choice based on several aspects: 1) It provides comparability across baselines and previous work, and 2) compared to large web document corpora, such as Common Crawl, it requires less data cleaning and is computationally more tractable, which improves the replicability of our results and helps to ensure that the major variable being evaluated is model performance (rather than engineering effort). Hence, while we believe that using a web-scale corpus, such as Common Crawl, would potentially enable even stronger baselines, we leave such experiments to future work.

\begin{table*}
\small

\resizebox{1\linewidth}{!}{
\begin{tabular}[t]{llll|c|cc|cc}
\toprule
\multirow{2}{*}{\textbf{Retriever}} & \multirow{2}{*}{\textbf{Reader}}  & \multicolumn{2}{c}{\textbf{Translation}} & \multicolumn{1}{c}{\textbf{Retrieval Metrics}} & \multicolumn{2}{c}{\textbf{Answerable Metrics}} & \multicolumn{2}{c}{\textbf{End-to-End Metrics}} \\
& & \textbf{Query} & \textbf{Answer} & \textbf{R@1} & \textbf{Mean}  $\mathbf{A}\in{}\mathbf{D}$ \textbf{F1} & \textbf{Mean} $\mathbf{A}\notin{}\mathbf{D}$ \textbf{F1} & \textbf{En F1} & \textbf{Mean F1} \\
\midrule
\textsc{No Answer} & - & - & - & - & - & - & 32.4 & 32.4 \\
\midrule
\multicolumn{9}{c}{\it MULTILINGUAL RETRIEVER} \\
\midrule
\textsc{Elasticsearch*} & \textsc{Xlm-R} & - & - & 42.57 \scriptsize{\textpm 1.2} & 25.18 \scriptsize{\textpm 3.8} & 7.24 \scriptsize{\textpm 2.5} & 34.99 & 34.13\scriptsize{ \textpm 0.4} \\
\midrule
\multicolumn{9}{c}{\it TRANSLATE-TEST ENGLISH RETRIEVER} \\
\midrule
\textsc{DPR} & \textsc{RoBERTa} & Test & Test & 53.62 \scriptsize{ \textpm 2.2} & 20.33 \scriptsize{\textpm 4.1}  & 10.24 \scriptsize{\textpm 1.8} & 45.19 & 36.81\scriptsize{ \textpm 1.2} \\
\midrule
\multicolumn{9}{c}{\it GOLD NQ PASSAGES} \\
\midrule
\textsc{Gold NQ} & \textsc{M-Bert} & - & \multirow{6}{*}{Test} & \multirow{6}{*}{80.22} & 20.13 \scriptsize{\textpm 5.5} & 7.56 \scriptsize{\textpm 1.7} &\multirow{3}{*}{51.97}& 37.8\scriptsize{ \textpm 2.0} \\
\textsc{Gold NQ} & \textsc{M-Bert} & Test &  &  & 28.10 \scriptsize{\textpm 6.5} & 12.1 \scriptsize{\textpm 2.1} & & 41.4\scriptsize{ \textpm 2.2} \\
\textsc{Gold NQ} & \textsc{M-Bert} & Train &  &  & \underline{32.21} \scriptsize{\textpm 6.0} & \underline{14.8} \scriptsize{\textpm 1.9} &  & \underline{44.1}\scriptsize{ \textpm 1.8} \\ \cline{6-9} \cline{1-3}
\textsc{Gold NQ} & \textsc{Xlm-R} & - & &  & 38.81 \scriptsize{\textpm 3.2} & 20.05 \scriptsize{\textpm 2.6} & \multirow{3}{*}{52.27} & 45.5\scriptsize{ \textpm 1.4} \\
\textsc{Gold NQ} & \textsc{Xlm-R} & Test &  &  &  34.23 \scriptsize{\textpm 5.0} & 16.38 \scriptsize{\textpm 2.6} & & 42.9\scriptsize{ \textpm 2.1} \\
\textsc{Gold NQ} & \textsc{Xlm-R} & Train &  &  &  \textbf{\underline{40.28}} \scriptsize{\textpm 3.1} & \textbf{\underline{20.93}} \scriptsize{\textpm 2.7} & & \textbf{\underline{46.0}}\scriptsize{ \textpm 1.4} \\
\midrule
\multicolumn{9}{c}{\it GENERATIVE MODELS} \\
\midrule
\textsc{Query-only} & \textsc{mT5} & - & - & - & - & - & 43.8 & 35.0\scriptsize{ \textpm 1.2} \\
\textsc{Gold NQ} & \textsc{mT5} & - & - & 80.22 & 36.8 \scriptsize{\textpm 6.2} & 17.07 \scriptsize{\textpm 2.6} & 47.6 & 38.5\scriptsize{ \textpm 2.2} \\
\bottomrule
\end{tabular}
}

\caption{\textbf{Results for each baseline, broken down by retrieval metrics (Recall @ K passages), answerable question metrics (F1 at the best confidence threshold) and end-to-end metrics (F1 at the best confidence threshold).} A naive approach, predicting exclusively \textsc{No Answer}, achieves a lower bound score of 32.42\% F1. Translate-Train using \gnq{}s Gold passages and an \textsc{Xlm-R} reader outperforms all alternate settings. \\
$\mathbf{A}\in{}\mathbf{D}$ \text{ denotes metrics for where the answer A exists in the top retrieved document D (exact match).} \\
$\mathbf{A}\notin{}\mathbf{D}$ \text{ denotes metrics for where the answer A does not exist in top retrieved document D (exact match).} \\
$^*$ Elasticsearch benchmark does not include Hebrew, Khmer, Korean, Malay, and Vietnamese.
\label{tab:baseline-results}}
\vspace*{-.4cm}
\end{table*}

\paragraph{\textbf{Elasticsearch $\rightarrow$ XLM-R}}
We benchmark a fully multilingual retriever approach using Elasticsearch followed by \textsc{Xlm-R} as the extractive reader.
Elasticsearch leverages language-specific tokenizers and analyzers with BM25 to search for native passages in the target language's Wikipedia dump. We used their built in language specific analyzers which include stopwords and stemmer in each language.\footnote{https://www.elastic.co/guide/en/elasticsearch/reference/\\
current/analysis-lang-analyzer.html\#arabic-analyzer}
We took the Wikipedia dump from Dec. 7, 2020 for each language as source documents.
The languages Hebrew, Khmer, Korean, Malay, and Vietnamese are not part of the Elasticsearch baseline as they are not natively supported by Elasticsearch.

\paragraph{\textbf{DPR $\rightarrow$ RoBERTa}}
We benchmark an approach that utilizes state-of-the-art English retrieval and reader systems, enabled by translating the incoming query into English, and the outgoing prediction into the target language.
We use off-the-shelf Dense Passage Retrieval  \citep[DPR,][]{karpukhin2020dense}, followed by \textsc{RoBERTa} \citep{liu2019roberta} to extract a prediction.\footnote{We use the trained ``Multiset" DPR model available in \url{https://github.com/facebookresearch/DPR}.}

\paragraph{\textbf{Gold NQ $\rightarrow$ Extractive QA}}
For this set of baselines, optimal English retrieval is simulated via the passages provided with \gnq{}.
We illustrate baselines that leverage these provided ``Gold" English documents, machine translation, and extractive QA models.
We vary the type of QA model (\textsc{M-Bert} vs.\ \textsc{Xlm-R}) and the train/test approach, comparing common \textbf{Zero Shot}, \textbf{Translate Test}, and \textbf{Translate Train} approaches.

In \textbf{zero shot} transfer each multilingual model is finetuned with \gnq{}s' default English questions $Q_{en}$ and passages $P_{en}$.
At test time the model receives MKQA questions $Q_{xx}$ in language $xx$, paired with English passages $P_{en}$.

For \textbf{translate test}, at train time the model uses \gnq{}'s default English. 
At test time, MKQA questions are translated into English $Q_{xx\rightarrow en}$, and the passage remains in English $P_{en}$.
Passages remain in English for both training and inference.

For \textbf{translate train}, at train time, questions are translated into the target language $Q_{en\rightarrow xx}$.
At test time the model is given queries in the target language $Q_{xx}$ and passages $P_{en}$ in the default English from \gnq{}.
Passages are always in English.

\respace
\paragraph{\textbf{Query-only mT5}}
We benchmark a ``closed-book", query-only generative QA approach, based on \citet{roberts2020much}.
This approach allows us to circumvent retrieval and machine translation entirely, using parametric knowledge within \textsc{mT5 Large}.
Simply, the query is fed to the model, which is trained to generate the localized answer directly.

\paragraph{\textbf{Gold NQ $\rightarrow$ mT5}}
We benchmark a stronger generative QA approach, that also has access to the English Gold NQ passages.
Based on open-source implementations for MLQA and XQuAD datasets, the model is fed the non-English query, with (in this case) the English gold passage, and generates the predicted answer.\footnote{Implementation and hyperparameters based on \url{https://github.com/google-research/multilingual-t5}.}

\respace
\subsection{Results}
\label{results}

Table~\ref{tab:baseline-results} presents retrieval and end-to-end metrics for each baseline, as the mean across all 26 languages.
Retrieval metrics include recall at K, measuring if the correct answer appears anywhere in the top K retrieved passages, as traditionally used in information retrieval settings.
Note that these metrics are computed by looking for an exact match of the text-normalized gold answer in the text-normalized passage.
We find that translation followed by English DPR outperforms the Elasticsearch multilingual sparse retrievers.
This is consistent with results observed in XOR-QA \citep{asai2020xor} which shows the surprising under-performance of multilingual retrievers.
Errors are likely a combination of no answer being present in smaller non-English Wikipedia indexes, and the weak performance of sparse retrieval.
The Gold \gnq{} documents contain a valid answer $80.22\%$ of the time.
However, this is likely an upper bound, as these documents are often very long and noisy, such that \gnq{} annotators often marked them as not containing an answer to the question, even though we find the gold answer string is present.

\begin{figure*}
\centering
\includegraphics[width=\linewidth]{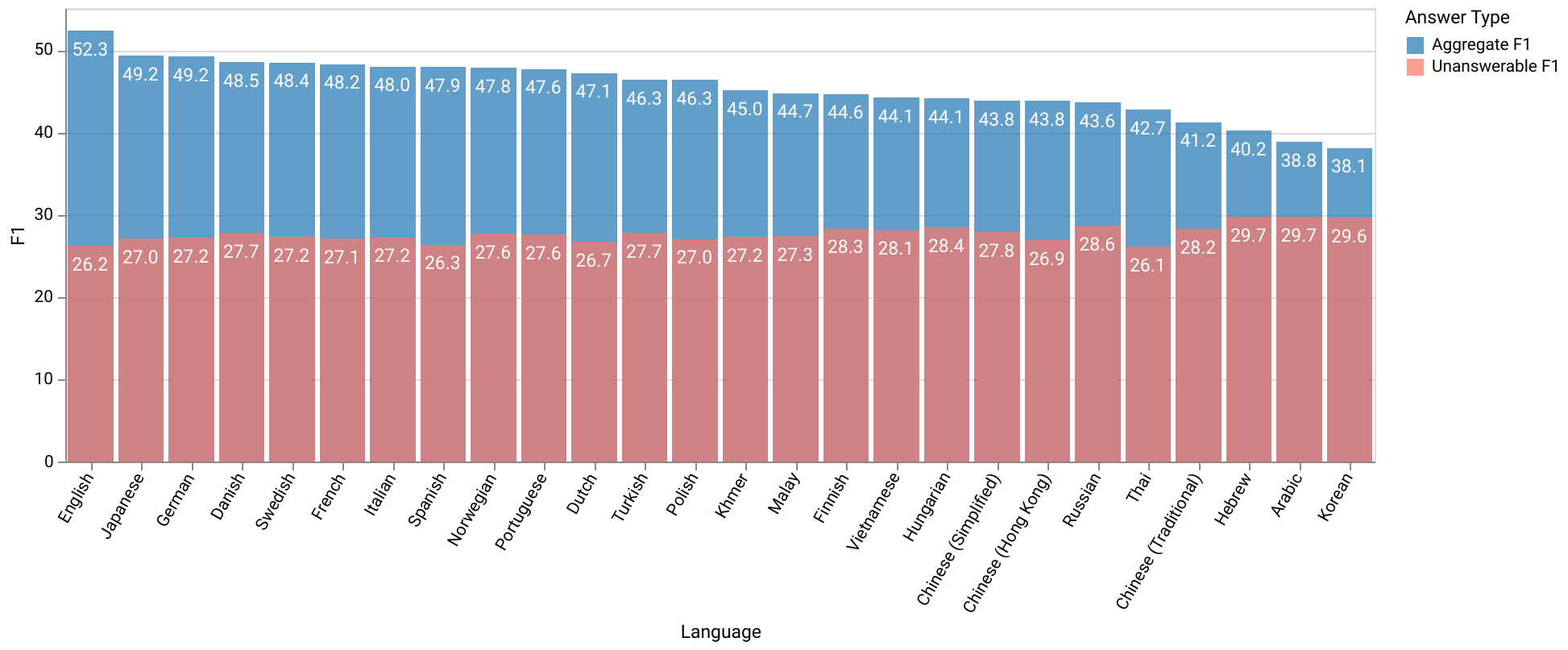}
\caption{\textbf{F1 by language.} \textsc{Xlm-R} Zero-Shot performance ranked by language. Unanswerable F1 (in red) corresponds to the proportion of the Aggregate F1 obtained from predicting No Answer. The Unanswerable proportion is calculated as the percentage of unanswerable examples ($32.42\%$) multiplied by the Unanswerable F1.}
\label{fig:language-results}

\end{figure*}
\respace

For end-to-end metrics, we measure F1 just for English (``EN F1"), which omits the impact of machine translation, and mean F1 over all 26 languages.
The naive baseline of only predicting No Answer achieves a lower bound score of 32.42\%.
We chose to combine both Unanswerable and Long Answers into the No Answer category for evaluation to focus \mkqa{} on short, factoid answers that can be evaluated automatically and robustly.
Unsurprisingly, we observe models with access to \gnq{} gold documents achieve the best results, with Translate Train \textsc{Xlm-R} achieving the best mean F1 of 46.0{\scriptsize $\pm$1.4}.
Among these methods, \textsc{Xlm-R} outperforms {M-Bert}, and Translate-Train outperforms Translate-Test and Zero Shot.
Generative approaches using \textsc{mT5} perform fairly well, even under zero shot conditions (trained only on English), or without any passage provided (query-only).

We also measure the F1 scores for the subset of answerable questions to measure the ability of the retrievers and readers to find the right answer. 
We separately report the average all-language F1 for (i) questions in which a gold answer appears in the top retrieved document, and (ii) questions in which none are found.
As expected, performance is much higher for both extractive and generative models where the retriever has succeeded. 
Translate Train with \textsc{Xlm-R} still achieves the best performance. 
\textsc{Xlm-R} also performs well on the correct outputs ($\mathbf{A}\in{}\mathbf{D}$) of the weakest retriever, Elasticsearch, though there are fewer of them.
Comparing with end-to-end metrics, which includes unanswerable questions, answerable questions are more difficult to answer.

Overall, these results show how collecting relevant passages remains a challenging bottleneck in multilingual open-retrieval QA.
Multilingual retrievers, English state-of-the-art retrievers, and generative QA models all fail to overcome this problem, and even when gold passages are provided, multilingual readers and machine translation still fail to consistently produce localized answers (with generous evaluation settings).

\respace
In Figure~\ref{fig:language-results} we compare cross-lingual performance between languages, ranked by F1 score.
We plot \textsc{Xlm-R} Zero Shot to minimize the noise from machine translation.
As expected, the \textsc{Xlm-R} model performs fairly well on English ($52.3$), and common non-English languages, including the most common Indo-European Germanic and Italic languages, but poorly on languages from lower-resourced families.
Note that the minimum F1 score is 32.42\%, where a threshold of $0$ predicts No Answer to every question. 
Interestingly, as the Aggregate F1 decreases, the Unanswerable F1 rises on average from ${\sim}27\%$ to ${\sim}29\%$, abstaining from an answer more often.
Given the parallel questions property of \mkqa{}, these metrics allow a practitioner to specifically identify languages with weak model performance, and answer abstention behaviour for commonly used reader models, such as \textsc{Xlm-R}.
Even before considering a cultural shift in query distribution, these metrics allow us to isolate performance on geographically invariant queries, and general effectiveness of transfer learning for particular languages and training regimes.

\subsection{Unanswerable vs.\ Long Answers}

As discussed in Section \ref{annotation-breakdown}, following the Short Answer setup for Natural Questions \citep{kwiatkowski2019natural} we define Unanswerable as a query without a short answer (i.e.\ examples with \textit{long} or \textit{unanswerable} answer types) --- for our task.
Although evaluating long answers is important, it is out of the scope of \mkqa{}.
The primary benefit of this decision is that it enforces the \textit{retrieval-independent annotations} property of \mkqa{}, since long answers have an unbounded number of correct answer strings.
Here we investigate whether \textit{long} and ``truly'' \textit{unanswerable} examples in \mkqa{} are treated differently by our baseline models.

To answer this question, we break down the larger Unanswerable set into the \textit{long} and `truly' \textit{unanswerable} examples, comprising 56\% and 44\% respectively.
We then compute the final performance (F1) by model type and by language for each of these two categories. 
We find the results vary according to the quality of the model and the language (as do performance on answerable queries), but the difference between the long answer and truly unanswerable scores are marginal. 
For instance, \textsc{Xlm-R} Translate Train, using Gold NQ passages, achieves 84.2\% F1 on \textit{long}, and 84.7\% on truly \textit{unanswerable} examples, with a mean difference over all 26 languages of only 0.5\%. 
These differences are similarly negligible across other baselines. 
This finding suggests standard open-domain QA systems, trained on short answer datasets like Natural Questions, have learned to consider long answers as unanswerable, and do not appear to find one set more challenging than the other.

\respace
\section{Discussion}
\respace
\label{discussion}
\paragraph{Difficulty of MKQA} Our baselines represent a strong and diverse set of methods, that score competitively with state-of-the-art on similar open domain question answering datasets.
Nonetheless, on English alone, the best system recieves an \textsc{F1} score of only 52.3\%, less than the same methods achieve on the open datasets Natural Questions and TriviaQA, or other standard benchmarks for this task.
These comparative results demonstrate \mkqa{} is highly challenging and leaves ample room for improvement in both English and the long tail of natural languages.
In this section we explain why, with a detailed comparison to its closest set, Natural Questions.

\begin{figure*}
\centering
\includegraphics[width=\linewidth]{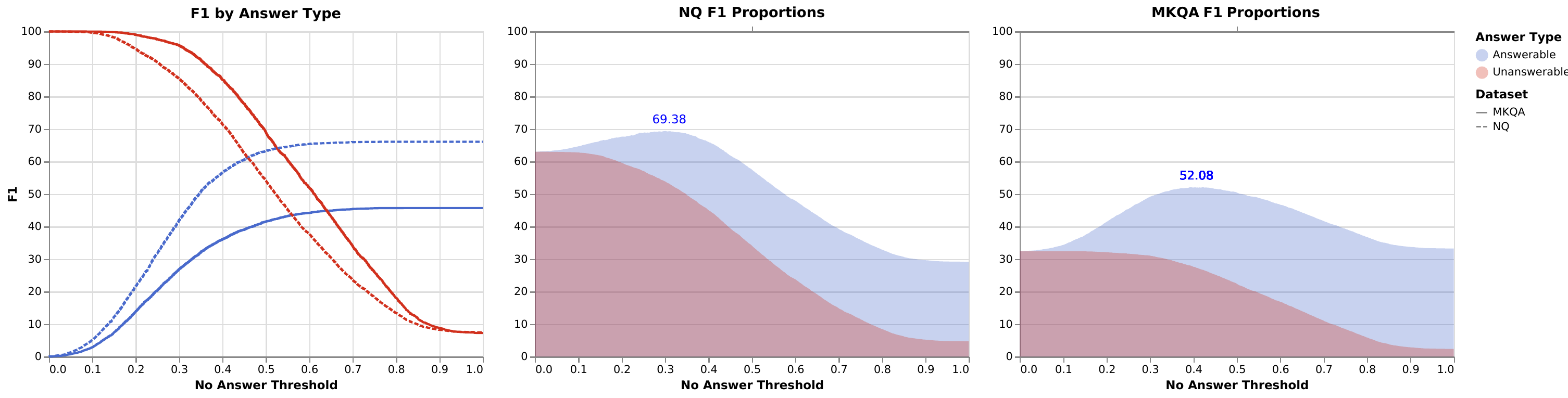}
\caption{\textbf{Comparing MKQA and \gnq{} English annotations.} The performance of the same English \textsc{Bert-Large} model on each of Natural Questions (\gnq{}) annotations and \mkqa{} annotations, using the \mkqa{} evaluation metrics. 
For all plots the y-axis is F1 score and the x-axis is the value of the threshold over No Answer probabilities. F1 by Answer Type (left diagram) compares the accuracy of the model on Answerable and Unanswerable examples for each dataset, showing Unanswerable examples are on average easier in \mkqa{}, and Answerable examples are on average harder in \mkqa{}. 
\gnq{} F1 Proportions (middle) and \mkqa{} F1 Proportions (right) show what proportion of the aggregate F1 score is derived from each Answer Type. 
These plots demonstrate \mkqa{} is more difficult than \gnq{} because there is a higher proportion of answerable questions, which are harder on average.}
\label{fig:gnq-vs-mkqa}
\vspace*{-.4cm}
\end{figure*}

Why is \mkqa{} so challenging for state-of-the-art approaches even for English open-domain QA?
To shed light on this, we compare the difficulty of English-only annotations between Natural Questions (\gnq{}) and \mkqa{}.
In Figure~\ref{fig:gnq-vs-mkqa} we use the same \textsc{Bert-Large} English model (trained on \gnq{}, using Gold \gnq{} passages) and evaluate it on both sets of annotations.
The ``F1 by Answer Type" diagram shows unanswerable examples in \mkqa{} (red line) are easier than the unanswerable examples in \gnq{} (red dashed line), as the model maintains higher performance at all No Answer confidence thresholds.
The opposite relationship is observed for answerable examples.

We hypothesize that this is due to the \textit{Retrieval-Independence} property and high coverage of our re-annotation process (described in \Cref{collection}).
Due to the annotation procedures \gnq{} uses, there are several cases that can lead to a potential answer missing from the dataset:
(a) the initial retrieval may have not produced a candidate, (b) the answer may have not been in Wikipedia, or (c) \gnq{} graders may have missed a valid answer.
\mkqa{} annotations are not susceptible to (a) and (b) and likely less impacted by (c).
Consequently, the most challenging questions migrated from unanswerable in \gnq{} to answerable in \mkqa{}, shifting the unanswerable distribution from 63\% to 32\% (as shown in Figure~\ref{fig:answer-type-breakdown}).
Consider the following examples.

\textbf{(a) \gnq{} retrieval failure} In this example, the \gnq{} retrieved document does not contain an answer to the question, causing no long or short answer (\textit{No Answer}) in \gnq{}.
There exists a better Wikipedia document (\href{https://en.wikipedia.org/wiki/Wheel_of_Fortune_(American_game_show)}{Wheel of Fortune}) that does contain the \mkqa{} answer ``Autumn Erhard''.

\respace
\begin{itemize}\itemsep0em
  \item \textbf{Q:} \textit{Who won the most money on wheel of fortune?}
  \item \textbf{NQ URL:} Wikipedia: \href{https://en.wikipedia.org//w/index.php?title=American_game_show_winnings_records}{\textit{American game show winnings records}}.
  \item \textbf{NQ Answer:} \textcolor{red}{\textit{No Answer}}
  \item \textbf{MKQA Answers:} \textcolor{PineGreen}{\textit{``Autumn Erhard''}}
\end{itemize}
\respace

\textbf{(b) No Wikipedia answer} This is also an answerable query, labelled as no answer by \gnq{}, because the answer is not found on Wikipedia (either by \gnq{} or our best efforts).
However, an answer can be found by MKQA graders from other websites and sources.

\respace
\begin{itemize}\itemsep0em
  \item \textbf{Q:} \textit{How many teeth does a saltwater crocodile have?}
  \item \textbf{NQ URL:} Wikipedia: \href{https://en.wikipedia.org//w/index.php?title=Saltwater_crocodile}{\textit{Saltwater Crocodile}}.
  \item \textbf{NQ Answer:} \textcolor{red}{\textit{No Answer}}
  \item \textbf{MKQA Answers:} \textcolor{PineGreen}{\textit{``66''}}
\end{itemize}
\respace

\textbf{(c) Annotator misses valid answer} For this query, the answer is clearly visible in the provided Wikipedia article, but \gnq{}'s annotation process yields no answer. 

\respace
\begin{itemize}\itemsep0em
  \item \textbf{Q:} \textit{What language do they speak in the ukraine?}
  \item \textbf{NQ URL:} Wikipedia: \href{https://en.wikipedia.org//w/index.php?title=Languages_of_Ukraine}{\textit{Languages of Ukraine}}.
  \item \textbf{NQ Answer:} \textcolor{red}{\textit{No Answer}}
  \item \textbf{MKQA Answers:} \textcolor{PineGreen}{\textit{``Ukrainian''}}
\end{itemize}
\respace

Given the answer to these queries are not easily found in the corpus, by retrieval, or by human annotators, they are likely more challenging on average.
As such, their label shift from no answer in \gnq{} to answerable in \mkqa{} likely explains why there is higher mean difficulty of answerable questions in \mkqa{}, as observed in Figure~\ref{fig:gnq-vs-mkqa}.
To understand the prevalence of each error type, we compute how often \textit{any} \mkqa{} answer appears in the retrieved document for which the \gnq{} label says no answer exists.
We find a valid answer appears in 70.4\% of these documents, suggesting category (c), annotator error, is the largest source of such unanswerable queries in \gnq{} (and the largest source of improvement in label quality for \mkqa{}).

The middle and right diagrams in Figure~\ref{fig:gnq-vs-mkqa} normalize the answer types by their proportion within the dataset, so we can compare their relative contributions to the aggregate F1 (the sum of answerable and unanswerable).
\gnq{} labels enable a much higher aggregate F1 score (69.38\% at the best threshold) than \mkqa{} (52.08\% at the best threshold) primarily due to the higher proportion of unanswerable examples --- which are easier on average than answerable examples.
By comparing the ratio of unanswerable to answerable examples attempted at the best thresholds in each of the middle and right diagrams (the blue regions vs.\ the red regions) we see that the \mkqa{} task is more oriented to answering questions rather than abstaining.

Due to the \textit{Parallel Question} property of \mkqa{}, the dataset is similarly challenging in all 26 languages.
There is also a noticeable gap between the performance on English and on lower-resourced languages (Figure~\ref{fig:language-results}).
For Korean and Arabic the best F1 score is only 6\% higher than the lower bound score of $32.42\%$ obtained from predicting exclusively ``unanswerable.''
This demonstrates that existing transfer learning methods have significant deficits to overcome for low-resource multilingual QA to match English performance.
\mkqa{} offers a challenging benchmark to measure this cross-language progress specifically.

\paragraph{Future Work} The parallel questions property of \mkqa{} offers alternative task setups in addition to typical open domain question answering.
\citet{lewis2019mlqa} suggests a generalized cross-lingual transfer task (G-XLT) where the question and answer languages are intentionally different.
Alternatively, future work might assume we are given the English question-answer pairs, and attempt to propagate these answers into other languages by localizing the questions and answers.

We anticipate that this dataset will enable industry practitioners and researchers to rapidly test and compare novel cutting-edge techniques for QA against existing techniques in a more fair, comparable, and precise manner than previous benchmarks.
Additionally, we hope that the linguistic diversity and large number of languages will inspire more researchers to treat model performance across many (partially less-resourced) languages as an important and worthy goal in itself.
As \mkqa{} offers the \textbf{only} open-QA option for many of these languages, we also hope to spark important research in these monolingual, non-English settings.

\respace
\section{Conclusion}
\respace

In this work, we introduce a multilingual open domain question answering evaluation set.
Its properties, including geographical invariance, language-parallel questions, retrieval-independent annotations, and linguistic diversity, set it apart from existing resources in terms of annotation quality, difficulty, and flexibility to evaluate new approaches.
We encourage future multilingual benchmarks to adopt data collection and annotation principles to promote higher-quality, and informative evaluation practices.
We evaluate several baselines, based on state-of-the-art methods, and demonstrate ample room for improvement both in English and in the tail of lower-resourced languages. 
We hope that this evaluation set enables wider exploration of cross-lingual and monolingual methods in non-English QA.

\section*{Acknowledgments}
We would like to thank Chris DuBois, who has been instrumental to releasing this data. Ilya Chatsviorkin, Xiao Ling, Nikhil Ramesh, Ni Lao, Agatha Downey, Silviana Ciurea-Ilcus, Anthony Chen, and Russ Webb have provided invaluable feedback on early versions of this paper. Thanks to Ivan Montero for testing out early versions of the data. Thanks to Pablo N.\ Mendes and Charles Srisuwananukorn for guidance and support, as well as to Noriyo Sakamoto for help in data collection. This work would not have been possible without the TryRating annotation platform.

\clearpage
\bibliography{tacl2018}
\bibliographystyle{acl_natbib}

\end{document}